\documentclass[letterpaper, 10 pt, conference]{ieeeconf}  

\IEEEoverridecommandlockouts                              

\usepackage[top=1in,bottom=.75in,right=0.75in,left=0.75in]{geometry}

\usepackage{cite}
\usepackage{amsmath,amssymb,amsfonts}
\newcommand{\norm}[1]{\left\lVert#1\right\rVert}
\usepackage[]{units}
\usepackage{array}
\usepackage{wasysym}
\usepackage{textcomp}
\usepackage{xcolor}
\usepackage{lipsum}
\usepackage{graphicx}
\usepackage{subfig}
\usepackage{caption} 
\usepackage[linesnumbered,ruled,vlined,english,onelanguage]{algorithm2e}
\SetAlFnt{\tiny}
\usepackage{booktabs}

\title{\LARGE \bf
Knowledge Transfer across Imaging Modalities Via Simultaneous Learning of Adaptive Autoencoders for High-Fidelity Mobile Robot Vision\\
}

\author{Md Mahmudur Rahman$^{1}$, Tauhidur Rahman$^{2}$, Donghyun Kim$^{2}$, and Mohammad Arif Ul Alam$^{1}$
\thanks{$^{1}$University of Massachusetts Lowell
        {\tt\small mdmahmudur\_rahman@student.uml.edu, mohammadariful\_alam@uml.edu}}%
\thanks{$^{2}$University of Massachusetts Amherst
        {\tt\small trahman@cs.umass.edu, donghyunkim@cs.umass.edu}}%
}

\begin{document}

\maketitle
\thispagestyle{empty}
\pagestyle{empty}

\maketitle

\begin{abstract}

Enabling mobile robots for solving challenging and diverse shape, texture, and motion related tasks with high fidelity vision requires the integration of novel multimodal imaging sensors and advanced fusion techniques. However, it is associated with high cost, power, hardware modification, and computing requirements which limit its scalability. In this paper, we propose a novel Simultaneously Learned Auto Encoder Domain Adaptation (\emph{SAEDA})-based transfer learning technique to empower noisy sensing with advanced sensor suite capabilities. In this regard, \emph{SAEDA} trains both source and target auto-encoders together on a single graph to obtain the domain invariant feature space between the source and target domains on simultaneously collected data. Then, it uses the domain invariant feature space to transfer knowledge between different signal modalities. The evaluation has been done on two collected datasets (LiDAR and Radar) and one existing dataset (LiDAR, Radar and Video) which provides a significant improvement in quadruped robot-based classification (home floor and human activity recognition) and regression (surface roughness estimation) problems. We also integrate our sensor suite and \emph{SAEDA} framework on two real-time systems (vacuum cleaning and Mini-Cheetah quadruped robots) for studying the feasibility and usability.
\end{abstract}


\section{Introduction}
Recent advancements of mobile robotic vision technologies (such as LiDAR, Camera, Depth, Infrared, Radar, Terahertz sensors) enable quadruped robots to navigate walking paths for blinds \cite{jo19}, play with owner \cite{YonezawaYUA08}, guardhouse from intruders \cite{zhang10}, remote monitoring of oil, gas and power installations and the construction sites\cite{asif20}. However, all of the above applications are involved a task-specific sensor suite which limits the scalability of the system \cite{jo19, YonezawaYUA08, zhang10, asif20}. Some of the sensors are more powerful than others in the task-specific applications, such as Lidars are generally found much superior to Radar in empowering mobile robots to solve versatile shape and movement sensing problems \cite{s20072068}. Recent advancements in radar imaging technology hold significant promise in versatile sensing applications with mobile robots due to affordability, reliability, and higher material penetrability (surface properties). But, radar still falls short in capturing precise and high-fidelity images of objects, surfaces, and human activities with a signal-to-noise ratio as high as Lidar \cite{s20072068}. To accommodate the capabilities of both of the sensors, recent quadruped robots use both imaging modalities and sometimes more advanced ones along with novel fusion techniques and algorithms facilitating significant improvement in robotic vision-based solutions \cite{lang2019pointpillars, ZhouT18, YangLU18}. Thus, scaling capabilities of existing mobile robots have become impossible without modifying existing hardware and algorithms, which forces customers to replace the old system with the latest one. In this paper, we present a way of improving the capabilities of existing quadruped robot vision, by utilizing a temporarily designed advanced sensor suite to align with existing sensors, collect few simultaneous data and use that to train a new algorithm to sustain the advanced sensor suite's capabilities without its presence i.e., empowering an existing mobile robot’s capabilities without any permanent hardware modifications.


Several recent works have explored simultaneous learning and knowledge transfer among multiple sensing modalities in solving complex robotic autonomy, navigation, and vision problems. Weston et. al. \cite{weston19} proposed a self-supervised deep learning-based method to estimate Radar sensor grid cell occupancy probabilities considering simultaneously collected Lidar-based probabilities as labels to train a deep regression model. \cite{kaul2020rssnet} developed a deep learning framework for Radar-based multi-object localization where the labels come from the sophisticated Lidar and RGB camera fusion-based models. In \cite{kim20}, researchers show that a Lidar-based localization framework ScanContext \cite{Kim2018ScanCE} can be used with radar data to improve the radar accuracy. An indoor localization using geometric structure was shown in \cite{ParkKK19} marking the radar points on lidar generated maps and CAD models. Yin et. al. \cite{yin2020radaronlidar} proposed a conditional generative adversarial network (GAN) based domain adaptation technique, the closest work to \emph{SAEDA} to generate Lidar representations of Frequency-Modulated Continuous-Wave (FMCW) Radar data and learn from it. A Monte Carlo localization (MCL) system is also formulated by motion and measurement models \cite{yin2020radaronlidar}. None of the above methods propose to utilize domain adaptation to enhance robotic vision capabilities aided by a temporarily advanced sensor suite and simultaneous learning schemes on temporarily collected simultaneous data. 

In this paper, we aim to answer two {\it {\bf key questions}: (q1) Can a temporarily integrated sensor suite collected high-dimensional heterogeneous simultaneous data be utilized to empower existing mobile robot vision with transfer learning? and (q2) Can we optimize the number of simultaneously collected labeled data i.e., semi-supervised learning?} Recent advancement of deep domain adaptation facilitates significant improvement in target domain classification performance in presence of label scarcity \cite{tzeng2017adversarial}, noise \cite{ csurka2017domain}, and heterogeneity \cite{tzeng2017adversarial} via generative networks (\cite{csurka2017domain, Tzeng_2015_ICCV, ganin2016domain}) or Discrepancy-based methods (\cite{sun2016deep, zellinger2017central, long2015learning,long16}). However, the generative networks-based domain adaptations training process is complex and does not guarantee convergence in the case of high-dimensional signals (Lidar/Radar) ({\it q1}) \cite{yao2019heterogeneous}. On the other hand, discrepancy-based methods related errors on the target domain are bounded by distribution divergence which makes it difficult to learn a domain invariant feature space between the heterogeneous source and target domain ({\it q1}) \cite{ben07}. Finally, all of the existing domain adaptation methods need an extensive amount of labeled source data that challenges {\it q2} requiring to design of a new domain adaptation algorithm for our target problem.

\begin{figure}
\begin{center}
  \includegraphics[width=\linewidth]{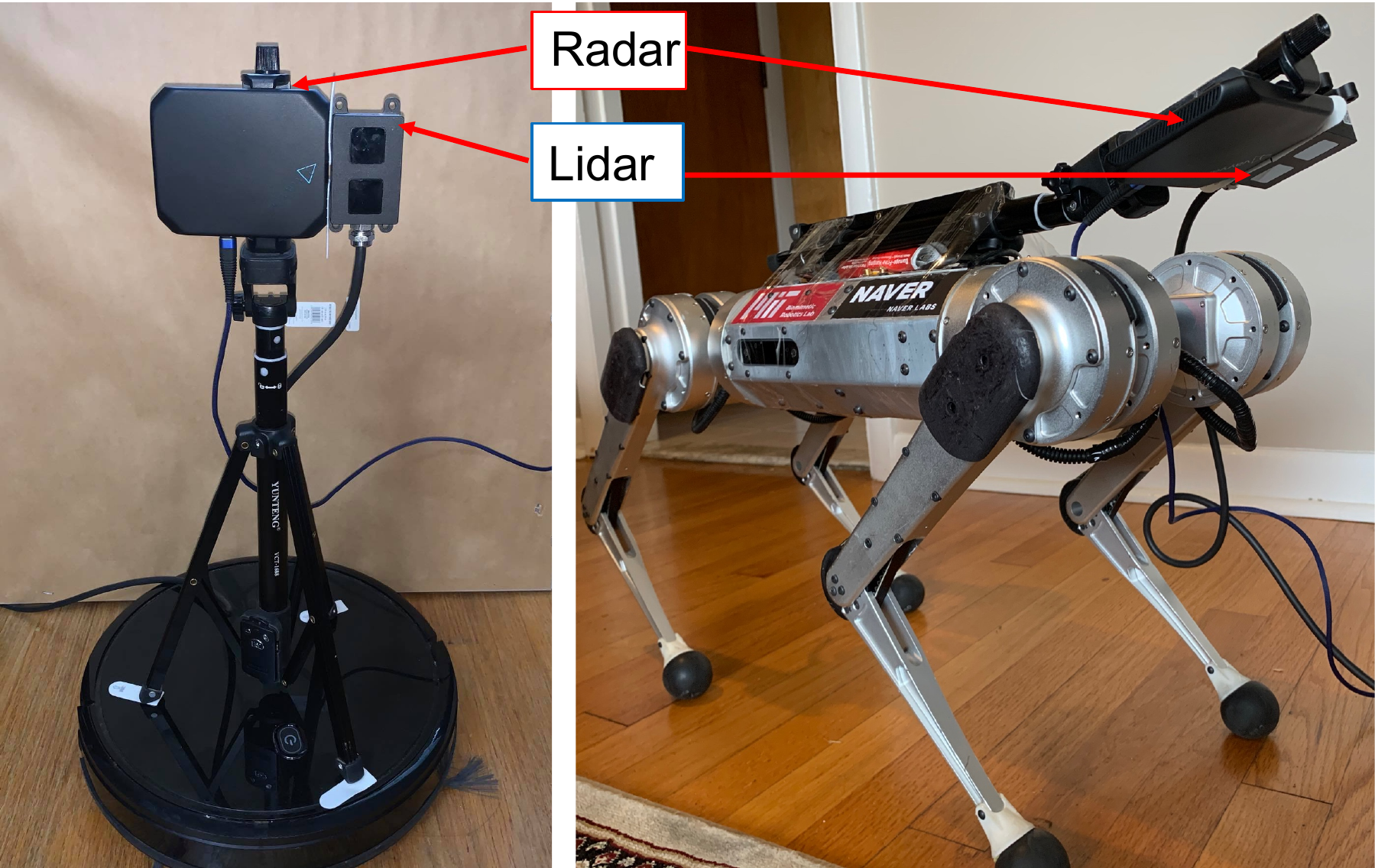}
  \caption{Our Lidar-Radar sensor suite integrated with Vacuum Cleaning Robot and Mini-Cheetah quadruped robot \cite{Katz2019MiniCA}}
  \label{fig:hardware}
\end{center}
\vspace{-5mm}
\end{figure}

In this paper, we propose \emph{SAEDA}, an auto-encoder based domain adaptation method which is both easy to train like discrepancy based methods \cite{long2017deep,zellinger2017central,sun2016return} and achieve a state of the art performance like adversarial networks \cite{tzeng2017adversarial, goodfellow2014generative} both in semi-supervised use cases \cite{long2017deep,zellinger2017central} in presence of a small amount of labeled target dataset. More specifically, our {\bf key contributions} are as follows:

\begin{itemize}
    \item We propose a novel simultaneous learning scheme for high dimensional imaging modalities (LiDAR/Radar) via a novel loss function which helps the target feature space to be adapted along the way of learning of the source domain feature space with only simultaneously collected data from existing imaging modality and temporarily designed high fidelity sensor suite.
    \item Additionally, we design a classification method by utilizing learned domain invariant feature spaces from the source and transferring them to the target to sustain characteristics of both discrepancy and adversarial based domain adaptation characteristics altogether for the semi-supervised scenario.
    \item We evaluate the performance of \emph{SAEDA} with real-time data on three use cases: (i) human activity recognition for robot-human interaction research, (ii) surface characterization to assist mobile robot for automated vacuum cleaning planning, and (iii) sandpaper surface estimation.
    \item Finally, we integrate our high fidelity sensor suite (LiDAR+Radar) in the Mini-Cheetah quadruped robot and study the efficacy/feasibility of our framework and system for outdoor surveillance.
\end{itemize}

\begin{figure*}
\begin{center}
  \includegraphics[width=\linewidth]{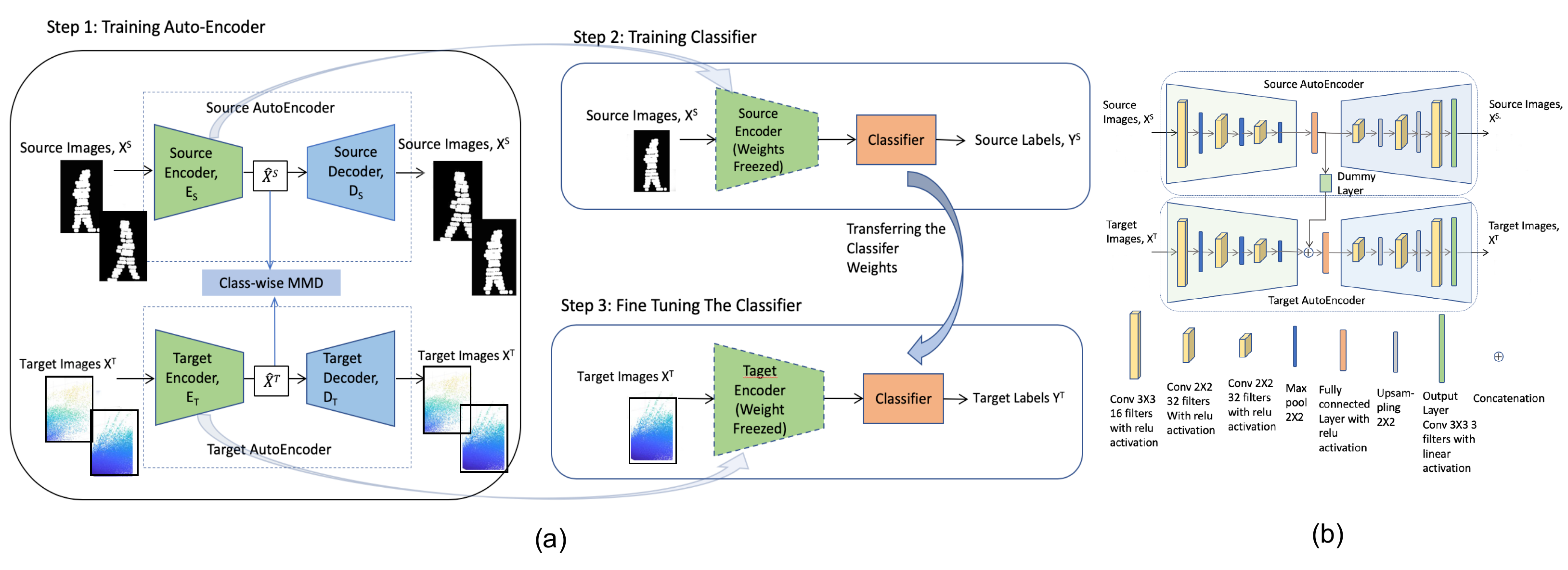}
  \caption{(a) Training procedures of our \emph{SAEDA} architecture. The dashed line represents the module is already trained and the weights are frozen in this step. The solid line represents the weights of this module are being updated in this step. (b) Network architecture of the source and the target auto-encoder.}
  \label{training_proc}
\end{center}
\vspace{-3mm}
\end{figure*}

\section{Deep Domain Adaptation Modeling}
\subsection{Setting}
We denote the source domain data having $n_s$ numbers of samples as $\mathcal{D}_s=\{(X_s^i, Y_s^i)\}^{n_s}$ where $X_s^i$ and $Y_s^i$ are the features and the class label of the $i^{th}$ sample respectively. Here, $X_s^i \in \mathbb{R}^{d_s}$; $d_s$ is the dimension of the source features. We define the source domain classification task, $T_s$ is to classify the source domain data correctly.
\par In case of the target domain data-set, we divide it into two dis-join sets, one for labeled ($\mathcal{D}_l$) and another for unlabeled ($\mathcal{D}_u$) samples. So, the target domain, $\mathcal{D}_t = \mathcal{D}_l \cup \mathcal{D}_u = \{(X_{tl}^{i}, Y_{tl}^{i})\}^{n_l} \cup \{(X_{tu}^{i}\}^{n_u}$. Here, $X_{tl}^{i}, X_{tu}^{i} \in \mathbb{R}^{d_t}$ are the labeled and the unlabeled $i^{th}$ target samples respectively where $d^t$ is the dimension of the target domain features. We assume $n_s \gg n_l$ and $n_u \gg n_l$, the number of target labeled sample is very smaller than number of target labeled and source labeled sample in our domain adaptation setup. Similar to $T_s$, the target classification task $T_t$ is defined to classify the unlabeled target data $X_{tu}^i$


\subsection{Adaptive Autoencoder via Simultaneous Learning}
Fig. \ref{training_proc} (a) shows the complete training process of \emph{SAEDA} where learning happens in three steps, simultaneous training of the auto-encoders, training the classifier, and fine-tuning.

\subsubsection{Training the Auto-encoders via Simultaneous Learning}

In this stage, the source and the target auto-encoders are trained with respective batches of feature vectors simultaneously. Eventually, the encoder parts of the auto-encoder, $E^s(\cdot)$ and $E^t(\cdot)$ map the input features $X_s$ and $X_t$ to the bottleneck feature representation space $\hat{X}_s$ and $\hat{X}_t$ respectively. 

\begin{equation}\label{eq:feature_mapping}
    \hat{X}_s = E^s(X_s), \quad \hat{X}_t = E^t(X_t)\\
\end{equation}

During the training, the objective function is set in a way that minimizes the statistical distance between $\hat{X}_s$ and $\hat{X}_t$. We propose to modify the Maximum Mean Discrepancy (MMD) \cite{gretton2006kernel} loss function and introduce the Class-wise MMD which minimizes the domain discrepancy according to the classes along the training. However, we apply a simultaneous learning scheme where both of the source and target auto-encoders get optimized simultaneously on a single graph developed on the top of both auto-encoders. This ensures matching of complex features between the source and the target domain along the way of convergence. We design a self reconstruction loss which sets the source auto-encoder optimization loss as follows.\\
$\mathcal{L}_s = - \frac{1}{N_s} \sum_{i=1}^{N_s} X_s^i \cdot log(p(X_s^i)) + (1-X_s^i) \cdot log(1-p(X_s^i))$

Here, $p(X_s^i)=E_s(D^s(X_s^i))$ is the source reconstruction probability of $X^s_i$ by the source auto-encoder with $N_s$ number of source sample.
On the other hand, we design a target auto-encoder loss function considering the weighted sum of self reconstruction loss and the class-wise MMD loss as follows.\\
    $\mathcal{L}_t = \mathcal{L}_{r} + \beta \cdot \mathcal{L}_{cws\_MMD}\;$\\
$\mathcal{L}_{r} =  - \frac{1}{N_t} \sum_{i=1}^{N_t} X_t^i \cdot log\;p(X_t^i) + (1-X_s^i) \cdot log\;(1-p(X_t^i))$

Here $\beta$ is a scaling parameter that determines the relative importance between the class-wise MMD loss ($\mathcal{L}_{cws\_MMD}$) and the reconstruction loss ($\mathcal{L}_{r}$).

\subsubsection{Class-wise MMD (cws-MMD) loss for Simultaneous Learning}
The MMD \cite{gretton2006kernel} loss function calculates the  statistical distance between the centroids of the source and the target feature distributions. So, The MMD loss function can be quantified as,\\
    $\mathcal{L}_{MMD}(\hat{X}_s, \hat{X}_t) = \norm{\frac{1}{n_s}\sum_{i=1}^{n_s} \hat{X}_s^i - \frac{1}{n_t}\sum_{i=1}^{n_t} \hat{X}_t^i }^2$

Where $\hat{X}_s^i$ and $\hat{X}_t^i$ are the $i^{th}$ feature representation for source and target data respectively.
\par In designing our cws-MMD loss, we intend to calculate the divergence between class conditional probability distribution $\mathcal{P}(X|Y)$ between the source and the target datasets. As a result, the divergence of class conditional probability can be approximated by calculating the distance between the centroids of corresponding source and the target class. Then, cws-MMD loss function can be defined as,

\begin{equation}\label{eq:invariant}
\begin{split}
   \mathcal{L}_{cws\_MMD} &(\hat{X}^s, \hat{X}^t) = \\
   &\frac{1}{C} \sum_{k=1}^C \norm{\frac{1}{n_s^k}\sum_{i=1}^{n_s^k} \hat{X}^s_{k,i} - \frac{1}{n_t^k}\sum_{i=1}^{n_t^k} \hat{X}^t_{k,i} }^2 
\end{split}
\vspace{-5mm}
\end{equation}

Here, $C$ is the maximum number of classes in the source and the target data. This new class-wise loss function aligns the source and target feature spaces according to the classes rather than the whole dataset blindly. This ensures the feature space alignment class by class for between the source and target domains.

\subsubsection{Training and fine-tuning the Classifier}
With the trained source and the target auto-encoders, we have a fine-tuned domain invariant feature space that can be used for semi-supervised (very few target labels) transfer learning. First, we train the classifier network with already trained feature space $\hat{X}_s$ from the source encoder as features and the labels of the source domain samples as the target. We then freeze the learned classifier network and use it to predict the classes of the target feature representation $\hat{X}_t$ in a semi-supervised fashion. The objective function of learning for the classifier network is,\\
$\min_{f_c} \mathcal{L}_c[Y^s, f_c(\hat{X}_s)]$

 Here, the classifier network is represented by $f_c(\cdot)$. We use categorical cross-entropy loss for optimization of classifier network. The complete training procedure of our proposed \emph{SAEDA} framework is presented in Algorithm \ref{algo_1}.
 

\begin{algorithm}
\tiny
\begin{small}
\SetAlgoLined
%
\SetKwInOut{Input}{Input}\SetKwInOut{Output}{Output}
\SetKwRepeat{Do}{do}{while}
\Input{Labeled Source Domain, $\mathcal{D}^s = \{X^s, Y^s\}$, Labeled Target Domain, $\mathcal{D}^t_l = \{X^t_l, Y^t_l\}$, Unlabeled Target Domain, $\mathcal{D}^t_u =  \{X^t_u\}$, model parameter $\beta$, number of classifier layers $c_l$, and bottleneck space size $b$}
\Output{Prediction class labels of unlabeled target domain, $\mathcal{D}^t_u =  \{X^t_u\}$}
Match the number of samples by class in $\mathcal{D}^s$ and $\mathcal{D}^t_l$ by randomly resampling the smaller class-domain\;
Sort $\mathcal{D}^s$ and $\mathcal{D}^t_l$ by class.
Initialize the source and the target auto-encoder weights randomly\;
Set the loss function $\mathcal{L}_s$ and $\mathcal{L}_t$ to the source and the target auto-encoder\;
\Repeat{$\mathcal{L}_s$ and $\mathcal{L}_t$ converges}{
Train source and target auto-encoders with $\{X^s, X^s\}$ and $\{X^t_l, X^t_l\}$ respectively. 
}

Take only Encoder part of source AE network and append Classifier network with it\;
Freeze the Encoder and randomly initialize the Classifier\;
\Repeat{Test Loss converge}{
Train Encoder + Classifier network with $X^s, Y^s$\;
}

Take target encoder and cascade with classifier network\;
Freeze target encoder part of the network\;
\Repeat{test loss converge}{
Train target encoder + classifier network with labelled target data, $\{X^t_l, Y^t_l\}$\;
}
Predict the label of  the target unlabelled target data, $\mathcal{D}^t_u =  \{X^t_u\}$ with encoder + classifier network\;

 \caption{Training Procedure of Auto-encoder based Domain Adaptation (SAEDA)}
 \label{algo_1}
\end{small}
\end{algorithm}

\subsection{Why SAEDA Works}
Considering our auto-encoder based framework depicted in Eq.~\eqref{eq:feature_mapping},\eqref{eq:invariant},  and Fig.~\ref{training_proc}, we can have the following assumptions:

\paragraph{Hypothesis 1}: The bottleneck feature distribution of the same classes in the source and the target domains are statistically similar. Formally, if $C_s = C_t$ then, $E_s(X^s_{C_s}) \cong E_t(X^t_{C_t})$ where $C_s$ and $C_t$ are the class label of source and target domain data respectively.
\paragraph{Hypothesis 2}: The bottleneck feature distribution of different classes are statistically dissimilar. If $C_s \neq C_t$ then, $E_s(X^s_{C_s}) \ncong E_t(X^t_{C_t})$.
\paragraph{Hypothesis 3}: Let denote $X_C$ is the set of feature samples of both source and target domain of class $C$. For each $C$, there exist a domain encoder $E^*(X_C)$ and a decoder $D^*(E^*(X_C))$, and
\begin{small}
    \begin{equation}
        \lim_{X_C \xrightarrow{}\infty} \frac{1}{X_C} \mathcal{L}_{cws\_MMD}(P_{\hat{X}_{s \xrightarrow{}t}}(\cdot|C,V), P_{X_C}(\cdot|C,V)) = 0
    \end{equation}
\end{small}
Where $P_{\hat{X}_{s \xrightarrow{}t}}(\cdot|C,V)$ is the conditional probability of feature embedding transfer from source to target with respect to the class $C$ and the domain invariant feature space $V$. Star (*) notation indicates to include both of the source and target domains.

\begin{figure}[!htb]
\begin{minipage}{0.23\textwidth}
\begin{center}
        \includegraphics[width=\textwidth]{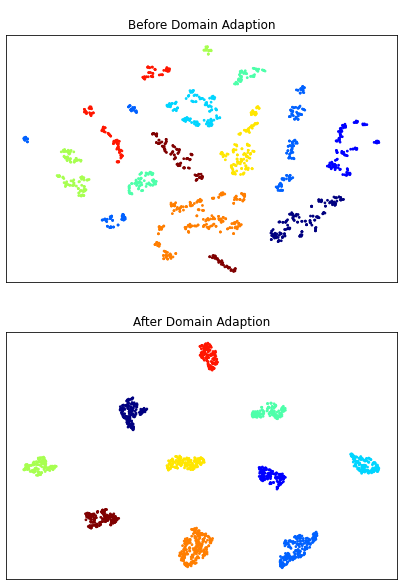} 
        \caption{The t-SNE visualization of the target domain feature representation space, before domain adaptation (top) and after domain adaptation (bottom). Different colors represent different classes of floor surface classification problem. } 
        \label{fig:t_sne} 
\end{center}
\end{minipage}
\begin{minipage}{0.23\textwidth}
\begin{center}
        \includegraphics[width=\textwidth, height=5cm]{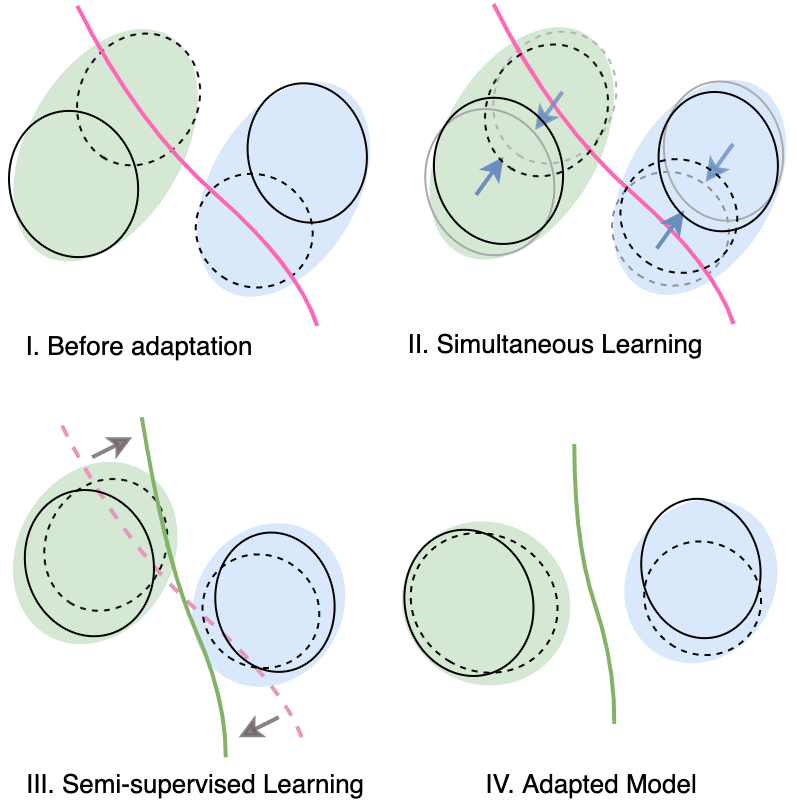} 
        \caption{The illustration of our SAEDA learning framework. Solid and dashed circles represent the source and the target domain respectively. The green and blue region represent two different classes.} 
        \label{fig:teaser} 
\end{center}
\end{minipage}
\end{figure}

If the bottleneck feature space dimension is set critically, the optimization loss function in Equations 3 and 4 will satisfy the ideal domain adaptation property in equation 9. This will match the class-wise domain distribution between the source and the target domain. In this case, it will only hold the domain invariant feature space representation $V$ for every class and discard the domain-specific feature space representations $U^s$ and $U^t$. This extraction of domain invariant information $V$ in the bottleneck layers enables efficient distribution matching between source and target domain in the corresponding classes. In Fig~\ref{fig:teaser}-I, the decision boundary (the solid red line) is only trained on the source data which easily violates the feature space of target data. With a simultaneous learning framework, the source and the target feature spaces align one with another (Fig~\ref{fig:teaser}-II). Semi-supervised fine-tuning with small numbers of target labeled data enables decision boundary to align according to the newly aligned source and the target feature spaces (Fig~\ref{fig:teaser}-III). Eventually, our SAEDA model generates a robust decision boundary (Fig~\ref{fig:teaser}-IV).

\section{Experimental Setup}

Our sensor suite consists of an Imaging Radar (Vayyar Imaging Radar Walabot 60 GHz) and a Solid-State Lidar (Hypersen Solid-State Lidar). We align both of the sensors together focusing on a single region of interest (ROI) as show in Fig. \ref{fig:hardware}. Table \ref{tab:lidar_radar_comparison} provides detailed comparisons between Lidar and Radar, that prove that Radar is much efficient in terms of cost, computation (lower resolution and Frame per seconds (FPS)), and power consumption. We integrate our prototype Lidar-Radar pair with Vacuum Cleaning Robot and Mini Cheetah via USB for both power supply and data streaming.

\begin{table}
  \caption{Vayyar Imaging Radar and Hypersen Solid State Lidar Comparisons.}
  \label{tab:lidar_radar_comparison}
  \centering
  \begin{tabular}{|l|l|l|}
  \hline
    {\bf Parameters} & {\bf Radar} & {\bf Lidar}\\
    \hline
    Cost & \$599 USD & \$699 USD\\
    \hline
    Resolution & 450$\times$60 & 320$\times$240\\
    \hline
    FPS& 8 & 17 \\
    \hline
    Power Supply & 4.5V & 9-12V \\
    \hline
    Current & 0.4-0.9 A & 0.25 - 1.2 A\\
    \hline
  \end{tabular}
  \vspace{-3mm}
\end{table}

\begin{table}[h]
\begin{center}
  \caption{Relation of sandpaper grit and surface roughness.}
  \label{tab:sand_paper}
  \begin{tabular}{|p{2cm}|p{0.7cm}|p{0.7cm}|p{0.7cm}|p{0.7cm}|p{0.7cm}|}
  \hline
    {\bf Sandpaper grit} & 120 & 240 & 320 & 500 & 1000 \\
    \hline
    {\bf Roughness ($\mu m$)} & 59.5 & 30.0 & 23.1 & 15.1 & 9.2 \\
    \hline
  \end{tabular}
\end{center}
\vspace{-5mm}
\end{table}

We collected two different datasets in the lab environment and used an existing dataset which is described below:
\begin{enumerate}
    \item {\bf Floor Dataset}: We integrated our Lidar and Radar sensor suite in an automatic vacuum cleaning robot (as shown in Fig. \ref{fig:hardware}) and collected simultaneous data on different floor types consists of (1) wooden floor, (2) tiles, (3) thin carpet, (4) thick carpet, (5) leather, (6) wet wooden floor, (7) wet tiles and (8) wet leather. The distance from the ground to the sensor is 36 cm. The data collection has been done for 5 minutes for each of the surface each, in a total, 40 minutes of collected data. We consider a 5 seconds window with 40\% overlap for classification.
    \item {\bf Sandpaper Dataset}: We used sandpaper with various grits as the surface roughness according to the ISO 6344 international standard. The used types of sandpaper used and their respective surface roughness as presented in Table \ref{tab:sand_paper}. To measure the surface roughness, a PMMA support was fixed firmly at a distance of 25 cm from the sensor surfaces, and sandpapers of different roughness were fixed to it. We performed the data collection 30 times by change angels, in total 30,000+ frames for each sensor (Lidar and Radar).
    \item {\bf LAMAR Human Activity Dataset}: This is an existing dataset consists of three imaging modalities (video camera, millimeter-wave Radar and Lidar) and 6 student volunteers (graduate, undergraduate and high school students) with $7$ different activities ("bending", "check\_watch", "call", "single\_wave",  "walking", "two\_wave", "normal\_standing") in 3 different rooms.\cite{alam20, benedek18}
\end{enumerate}

\subsection{Implementation Details}
We implement \emph{SAEDA} with TensorFlow and Keras. We keep the data pre-processing and the batch size consistent all over the data-sets. We keep the architecture of the auto-encoders symmetric along with the bottleneck layer as shown in Fig \ref{training_proc}(b). As a result, the topology of decoder layers is a reverse of the encoder layers. We use upsampling $2\times2$ layers in the decoder modules instead of maxpool $2\times2$ layer in the encoder modules to keep the symmetry. There are three convolution layers and one fully connected layer in every encoder and the decoder module. We use the convolution layer as the first layer of the encoder and the second to the last layer of the decoder having 16 filters and a filter size of $2\times2$. In case of the second layer of the encoder and the third to the last layer of the decoder, we use another convolution layer having 32 numbers of filters and a filter size of $3\times3$. The bottleneck layer consists of 100 neurons with relu activation. Finally, we use an output convolution layer as the last layer of the decoder module of filter size $3\times3$ and the number of filters same as the number of channels in the images. We also use a dummy layer between the bottleneck layers to maintain the source and target auto-encoders in a single graph so that we can train both simultaneously (as shown in Fig \ref{training_proc}(b)). The dummy layer consists of a unity weight with zero activation function so that it does not affect any functionality of our system. We use Adam \cite{kingma2014adam} optimization function with learning rate of $1\times10^{-4}$ to optimize the network. We optimize the learning rate and the parameter $\beta$ using hyper-parameter tuning. The final value of $\beta$ is 0.25. We run our \emph{SAEDA} model on a server having Nvidia GTX GeForce Titan X GPU and Intel Xeon CPU (2.00GHz) processor with 12 Gigabytes of RAM.
\begin{figure*}[h]
\begin{center}
  \centering
  \includegraphics[width=\linewidth]{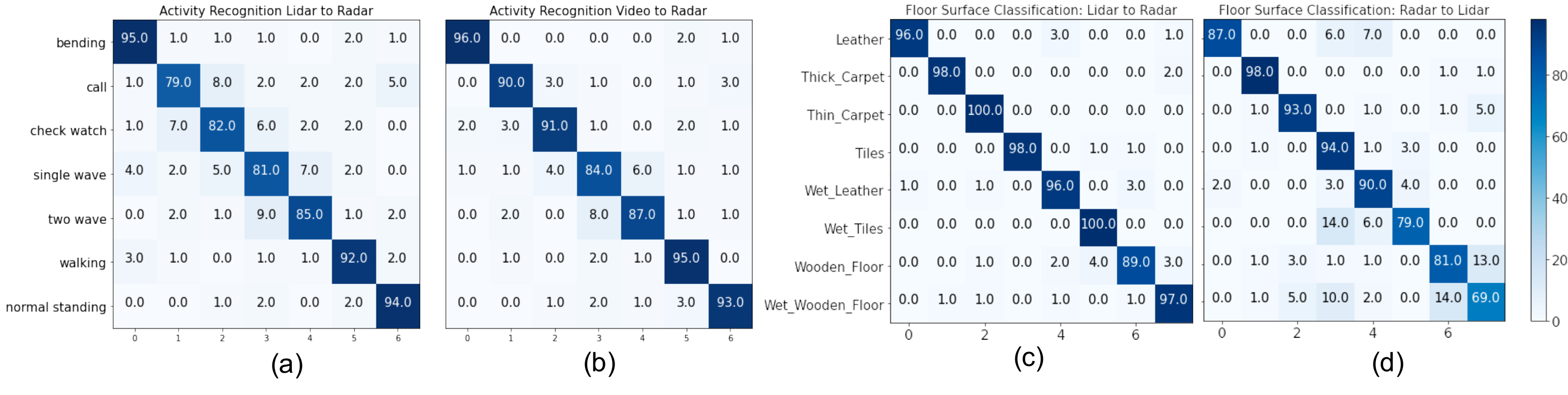}
  \vspace{-5mm}
  \caption{Confusion Matrix of activity recognition accuracy for (a) Lidar $\rightarrow$ Radar and (b)  Video $\rightarrow$ Radar domain adaptation  on LAMAR dataset as well as floor surface classification for (c) Lidar $\rightarrow$ Radar and (d) Radar $\rightarrow$ Lidar on our dataset}
  \label{fig:confusion_matrix}
  \vspace{-5mm}
\end{center}

\end{figure*}

\section{Results}

\begin{figure}
\begin{center}
  \centering
  \includegraphics[width=\linewidth]{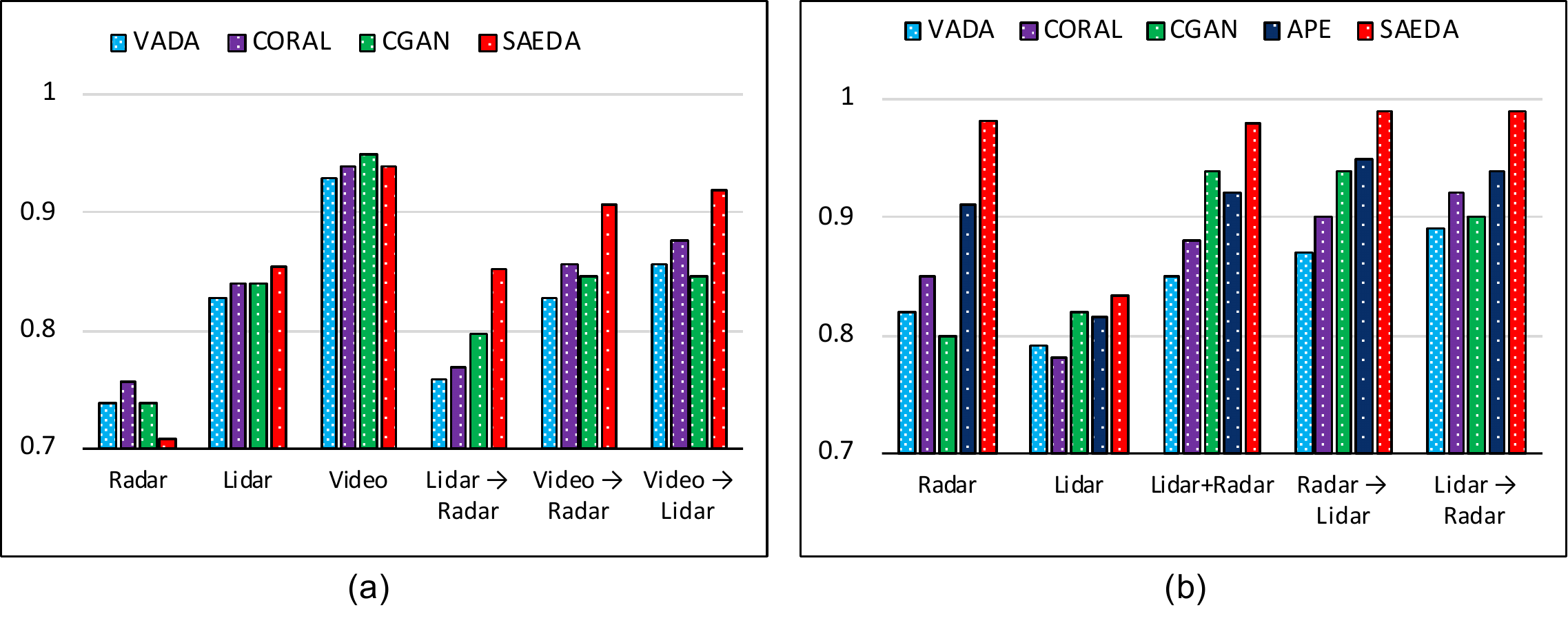}
  \caption{(a) Activity Recognition Result Comparisons with Baseline (b) Floor surface recognition result Comparisons with Baseline methods}
  \label{fig:classification_accuracy}
\end{center}
\vspace{-5mm}
\end{figure}

We implement {\bf four baseline methods}: Deep Correlation Alignment (CORAL) \cite{DBLP:journals/corr/SunS16a}, Virtual Adversarial Domain Adaptation (VADA) \cite{shu2018dirtt}, Conditional GAN (CGAN) \cite{ParkKK19} and Attract, Perturb, and Explore (APE) \cite{kim2020attract}.

\subsection{Classification Results: Activity and Floor Recognition}
Fig.\ref{fig:classification_accuracy}(a) and Fig. \ref{fig:classification_accuracy}(b) show performance comparisons of \emph{SAEDA} framework with baseline methods. It can be depicted that \emph{SAEDA} outperforms baseline methods significantly providing on average 4.5\% and 3.2\% improvements of accuracies than the nearest baseline methods for activity recognition and floor surface recognition respectively. Fig. \ref{fig:confusion_matrix} shows the confusion matrix of activity recognition and floor surface recognition for both Lidar $\rightarrow$ Radar and Radar $\rightarrow$ Lidar domain adaptations. Fig. \ref{fig:confusion_matrix} shows that Radar $\rightarrow$ Lidar provides extremely low accuracy comparing to Lidar $\rightarrow$ Radar adaptations. The accuracy differences are expected as we know 60 GHz Radar is highly absorbed by water providing significant changes in signals while Lidar signals are partially absorbed by water. On the other hand, Radar $\rightarrow$ Lidar provides better accuracy than Lidar $\rightarrow$ Radar due to Lidar's higher accuracy in position measurement (precision level is 1 cm for Hypersen Solid-State Lidar) than Radar.
\begin{table}[t]
  \caption{Surface Roughness Estimation Result comparisons on Sandpaper dataset. Surface roughness has been estimated as (Microns) $\mu m$. $R^2$ score has been used to calculate the goodness of the estimation performance}
  \label{tab:surface_roughness}
 \centering
 \begin{tabular}{p{1.6cm} p{0.9cm} p{0.9cm} p{0.9cm} p{0.9cm} p{0.9cm}}
 \toprule
    Method &  Radar & Lidar & Radar $\rightarrow $Lidar  & Lidar$ \rightarrow$ Radar & Radar + Lidar   \\
    \midrule
   CORAL \cite{long2015learning} & $0.8350$ & $0.8532$ & $0.9021$ & $0.9143$ & $0.9235$\\
   VADA \cite{shu2018dirtt} & $0.8135$ & $0.8298$ & $0.9147$ & $0.9276$ & $0.9421$\\
    CGAN \cite{ganin2016domain} & $0.7064$ & $0.7673$ & $0.7661$ & $0.9024$ & $0.9143$\\
    APE \cite{kim2020attract} & $0.9153$ & $0.9312$ & $0.9624$ & $0.9701$ & $0.9694$ \\
   \textbf{SAEDA} & $\mathbf{0.9848}$ & $\mathbf{0.8995}$ & $\mathbf{0.9976}$ & $\mathbf{0.9972}$ & $\mathbf{0.9986}$\\
   \bottomrule
  \end{tabular}
  \vspace{-5mm}
\end{table}


\subsection{Regression Results: Surface Roughness Characterization}
We substitute softmax layer of \emph{SAEDA} framework to a linear regression layer to create a regression problem for surface roughness estimation. Table \ref{tab:surface_roughness} shows the comparisons of $R^2 (R-Squared)$ and Mean Squared Error (MSE) metrics of the different algorithms in estimating surface roughness. R-squared quantifies the relative percentage measure of the variance of dependent variables from the model, which can be defined as $R^2 = 1 - \frac{\text{sum of squares of residuals}}{\text{total sum of squares}}$. It can be easily viewed that the Radar sensor outperforms Lidar in detecting surface roughness. Also, \emph{SAEDA} framework improves the domain adaptation significantly than the baseline methods.

\subsection{Outdoor Test using Mini-Cheetah Robot}
\begin{figure}[th]
\begin{center}
  \centering
  \includegraphics[width=\linewidth]{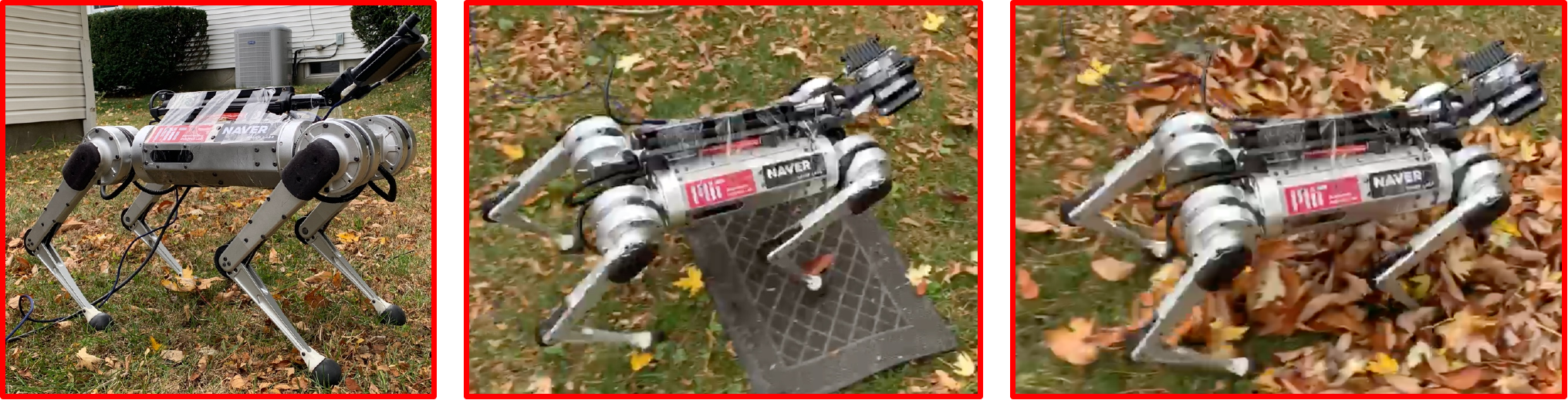}
  \caption{Mini-Cheetah walks over different surfaces (grass, carpet, leaf surfaces) while carrying our sensor suite.}
  \label{fig:mini_cheetah}
\end{center}
\vspace{-0.5mm}
\end{figure}

To see the feasibility of our roughness estimator in the four-legged robot application, we integrate our Radar and Lidar sensor suite with Mini-Cheetah and run an outdoor experiment. In this test, Mini-Cheetah moves around different surfaces which include thick grass, leather carpet, leaf, and thin grass (Fig. \ref{fig:mini_cheetah}). The locomotion controller of \cite{kim2019highly} is used and there is no feedback from the vision sensors in this test. We intentionally decouple the walking control from the estimation to watch that the terrain roughness is reasonably estimated even under the notable disturbance coming from the general walking behavior. The test results are summarized in Table \ref{tab:mini_cheetah_test}. It is visible that leaves are the softest surface in the wild while thick grass can be considered as the roughest surface as per our feasibility study.
\section{Discussion}
Figure \ref{fig:t_sne} shows the t-SNE\cite{van2008visualizing} feature representation of target domain features before and after domain adaptation. Different colors represent different classes in the target domain. From the t-SNE plot, it is evident that feature representation space is not well clustered before domain adaptation which leads to misclassification and hence lower classification score. On the other hand, the decision boundary on the well-clustered feature representation space can easily separate the classes correctly which leads to a higher classification score.

\begin{table}
\begin{center}
  \caption{Mini-Cheetah Robot Outdoor Test Result.}
  \label{tab:mini_cheetah_test}
  \begin{tabular}{|p{1.2cm}|p{1cm}|p{1.5cm}|p{1.6cm}|p{1cm}|}
  \hline
    {\bf Surface} & {\bf Grass} & {\bf ThickCarpet}& {\bf LeatherCarpet}& {\bf Leaves}\\
    \hline
    {\bf Roughness ($\mu m$)}  & $30.5\pm3.5$ & $19.5\pm2.3$ & $15.2\pm 2.6$ & $10.6\pm1.4$\\
    \hline
  \end{tabular}
\end{center}
\vspace{-0.8cm}
\end{table}

\section{Conclusion}
\emph{SAEDA} is capable of learning domain invariant feature space between simultaneously collected noisy and error-free sensor signals, enabling the development of the semi-supervised transfer learning framework. Experimental evaluation on three distinct recognition tasks using collected and public datasets, and feasibility study provides ample evidence that \emph{SAEDA} is capable of amplifying any robotic sensing performance with the involvement of least possible hardware and computational complexity. This framework, revolutionizing the idea of hardware integration virtualization, is able to provide existing deployed sensors-integrated autonomous system such capabilities that only could be achieved using expensive hardware modifications and computationally complex infrastructure or replacement of the latest version of mobile robotic systems.

\bibliographystyle{IEEEtran}  
\bibliography{reference} 

\begin{thebibliography}{10}
\providecommand{\url}[1]{#1}
\csname url@rmstyle\endcsname
\providecommand{\newblock}{\relax}
\providecommand{\bibinfo}[2]{#2}
\providecommand\BIBentrySTDinterwordspacing{\spaceskip=0pt\relax}
\providecommand\BIBentryALTinterwordstretchfactor{4}
\providecommand\BIBentryALTinterwordspacing{\spaceskip=\fontdimen2\font plus
\BIBentryALTinterwordstretchfactor\fontdimen3\font minus
  \fontdimen4\font\relax}
\providecommand\BIBforeignlanguage[2]{{%
\expandafter\ifx\csname l@#1\endcsname\relax
\typeout{** WARNING: IEEEtran.bst: No hyphenation pattern has been}%
\typeout{** loaded for the language `#1'. Using the pattern for}%
\typeout{** the default language instead.}%
\else
\language=\csname l@#1\endcsname
\fi
#2}}

\bibitem{jo19}
J.~a. Guerreiro, D.~Sato, S.~Asakawa, H.~Dong, K.~M. Kitani, and C.~Asakawa,
  ``Cabot: Designing and evaluating an autonomous navigation robot for blind
  people,'' in \emph{The 21st International ACM SIGACCESS Conference on
  Computers and Accessibility}, ser. ASSETS '19.\hskip 1em plus 0.5em minus
  0.4em\relax New York, NY, USA: Association for Computing Machinery, 2019, p.
  68–82.

\bibitem{YonezawaYUA08}
\BIBentryALTinterwordspacing
T.~Yonezawa, H.~Yamazoe, A.~Utsumi, and S.~Abe, ``Gazeroboard:
  Gaze-communicative guide system in daily life on stuffed-toy robot with
  interactive display board,'' in \emph{2008 {IEEE/RSJ} International
  Conference on Intelligent Robots and Systems, September 22-26, 2008,
  Acropolis Convention Center, Nice, France}.\hskip 1em plus 0.5em minus
  0.4em\relax {IEEE}, 2008, pp. 1204--1209. [Online]. Available:
  \url{https://doi.org/10.1109/IROS.2008.4650692}
\BIBentrySTDinterwordspacing

\bibitem{zhang10}
Y.~{Zhang} and Y.~{Meng}, ``A decentralized multi-robot system for intruder
  detection in security defense,'' in \emph{2010 IEEE/RSJ International
  Conference on Intelligent Robots and Systems}, 2010, pp. 5563--5568.

\bibitem{asif20}
M.~A. Arain, V.~H. Bennetts, E.~Schaffernicht, and A.~J. Lilienthal, ``Sniffing
  out fugitive methane emissions: autonomous remote gas inspection with a
  mobile robot,'' \emph{The International Journal of Robotics Research},
  vol.~0, no.~0, p. 0278364920954907, 2020.

\bibitem{s20072068}
\BIBentryALTinterwordspacing
C.~Debeunne and D.~Vivet, ``A review of visual-lidar fusion based simultaneous
  localization and mapping,'' \emph{Sensors}, vol.~20, no.~7, 2020. [Online].
  Available: \url{https://www.mdpi.com/1424-8220/20/7/2068}
\BIBentrySTDinterwordspacing

\bibitem{lang2019pointpillars}
H.~A. Lang, S.~Vora, H.~Caesar, L.~Zhou, J.~Yang, and O.~Beijbom,
  ``Pointpillars: Fast encoders for object detection from point clouds,''
  \emph{CVPR}, pp. 12\,697--12\,705, 2019.

\bibitem{ZhouT18}
Y.~Zhou and O.~Tuzel, ``Voxelnet: End-to-end learning for point cloud based 3d
  object detection,'' in \emph{2018 {IEEE} Conference on Computer Vision and
  Pattern Recognition, {CVPR} 2018, Salt Lake City, UT, USA, June 18-22,
  2018}.\hskip 1em plus 0.5em minus 0.4em\relax {IEEE} Computer Society, 2018,
  pp. 4490--4499.

\bibitem{YangLU18}
B.~Yang, W.~Luo, and R.~Urtasun, ``{PIXOR:} real-time 3d object detection from
  point clouds,'' in \emph{2018 {IEEE} Conference on Computer Vision and
  Pattern Recognition, {CVPR} 2018, Salt Lake City, UT, USA, June 18-22,
  2018}.\hskip 1em plus 0.5em minus 0.4em\relax {IEEE} Computer Society, 2018,
  pp. 7652--7660.

\bibitem{weston19}
R.~{Weston}, S.~{Cen}, P.~{Newman}, and I.~{Posner}, ``Probably unknown: Deep
  inverse sensor modelling radar,'' in \emph{2019 International Conference on
  Robotics and Automation (ICRA)}, 2019, pp. 5446--5452.

\bibitem{kaul2020rssnet}
P.~Kaul, D.~De~Martini, M.~Gadd, and P.~Newman, ``Rss-net: Weakly-supervised
  multi-class semantic segmentation with fmcw radar,'' \emph{arXiv preprint
  arXiv:2004.03451}, 2020.

\bibitem{kim20}
G.~{Kim}, Y.~S. {Park}, Y.~{Cho}, J.~{Jeong}, and A.~{Kim}, ``Mulran:
  Multimodal range dataset for urban place recognition,'' in \emph{2020 IEEE
  International Conference on Robotics and Automation (ICRA)}, 2020, pp.
  6246--6253.

\bibitem{Kim2018ScanCE}
G.~Kim and A.~Kim, ``Scan context: Egocentric spatial descriptor for place
  recognition within 3d point cloud map,'' in \emph{2018 IEEE/RSJ International
  Conference on Intelligent Robots and Systems (IROS)}, 2018, pp. 4802--4809.

\bibitem{ParkKK19}
Y.~S. Park, J.~Kim, and A.~Kim, ``Radar localization and mapping for indoor
  disaster environments via multi-modal registration to prior lidar map,'' in
  \emph{2019 {IEEE/RSJ} International Conference on Intelligent Robots and
  Systems, {IROS} 2019, Macau, SAR, China, November 3-8, 2019}.\hskip 1em plus
  0.5em minus 0.4em\relax {IEEE}, 2019, pp. 1307--1314.

\bibitem{yin2020radaronlidar}
H.~Yin, Y.~Wang, L.~Tang, and R.~Xiong, ``Radar-on-lidar: metric radar
  localization on prior lidar maps,'' 2020.

\bibitem{tzeng2017adversarial}
E.~Tzeng, J.~Hoffman, K.~Saenko, and T.~Darrell, ``Adversarial discriminative
  domain adaptation,'' in \emph{Proceedings of the IEEE Conference on Computer
  Vision and Pattern Recognition}, 2017, pp. 7167--7176.

\bibitem{csurka2017domain}
G.~Csurka, ``Domain adaptation for visual applications: A comprehensive
  survey,'' \emph{arXiv:1702.05374, arXiv preprint}, 2017.

\bibitem{Tzeng_2015_ICCV}
E.~Tzeng, J.~Hoffman, T.~Darrell, and K.~Saenko, ``Simultaneous deep transfer
  across domains and tasks,'' in \emph{The IEEE International Conference on
  Computer Vision (ICCV)}, December 2015.

\bibitem{ganin2016domain}
Y.~Ganin, E.~Ustinova, H.~Ajakan, P.~Germain, H.~Larochelle, F.~Laviolette,
  M.~Marchand, and V.~Lempitsky, ``Domain-adversarial training of neural
  networks,'' \emph{The Journal of Machine Learning Research}, vol.~17, no.~1,
  pp. 2096--2030, 2016.

\bibitem{sun2016deep}
B.~Sun and K.~Saenko, ``Deep coral: Correlation alignment for deep domain
  adaptation,'' in \emph{European conference on computer vision}.\hskip 1em
  plus 0.5em minus 0.4em\relax Springer, 2016, pp. 443--450.

\bibitem{zellinger2017central}
W.~Zellinger, T.~Grubinger, E.~Lughofer, T.~Natschl{\"a}ger, and
  S.~Saminger-Platz, ``Central moment discrepancy (cmd) for domain-invariant
  representation learning,'' \emph{arXiv preprint arXiv:1702.08811}, 2017.

\bibitem{long2015learning}
M.~Long, Y.~Cao, J.~Wang, and M.~I. Jordan, ``Learning transferable features
  with deep adaptation networks,'' \emph{arXiv preprint arXiv:1502.02791},
  2015.

\bibitem{long16}
M.~Long, H.~Zhu, J.~Wang, and M.~I. Jordan, ``Unsupervised domain adaptation
  with residual transfer networks,'' in \emph{NIPS}, 2016, pp. 136--144.

\bibitem{yao2019heterogeneous}
Y.~Yao, Y.~Zhang, X.~Li, and Y.~Ye, ``Heterogeneous domain adaptation via soft
  transfer network,'' in \emph{Proceedings of the 27th ACM International
  Conference on Multimedia}, 2019, pp. 1578--1586.

\bibitem{ben07}
J.~Blitzer, K.~Crammer, A.~Kulesza, F.~Pereira, and J.~Wortman, ``Learning
  bounds for domain adaptation,'' in \emph{NIPS}, 2008, pp. 129--136.

\bibitem{Katz2019MiniCA}
B.~Katz, J.~D. Carlo, and S.~Kim, ``Mini cheetah: A platform for pushing the
  limits of dynamic quadruped control,'' 2019, pp. 6295--6301.

\bibitem{long2017deep}
M.~Long, H.~Zhu, J.~Wang, and M.~I. Jordan, ``Deep transfer learning with joint
  adaptation networks,'' in \emph{Proceedings of the 34th International
  Conference on Machine Learning-Volume 70}.\hskip 1em plus 0.5em minus
  0.4em\relax JMLR. org, 2017, pp. 2208--2217.

\bibitem{sun2016return}
B.~Sun, J.~Feng, and K.~Saenko, ``Return of frustratingly easy domain
  adaptation,'' in \emph{Thirtieth AAAI Conference on Artificial Intelligence},
  2016.

\bibitem{goodfellow2014generative}
I.~Goodfellow, J.~Pouget-Abadie, M.~Mirza, B.~Xu, D.~Warde-Farley, S.~Ozair,
  A.~Courville, and Y.~Bengio, ``Generative adversarial nets,'' in
  \emph{Advances in neural information processing systems}, 2014, pp.
  2672--2680.

\bibitem{gretton2006kernel}
A.~Gretton, K.~Borgwardt, M.~Rasch, B.~Sch{\"o}lkopf, and A.~Smola, ``A kernel
  method for the two-sample-problem,'' \emph{Advances in neural information
  processing systems}, vol.~19, pp. 513--520, 2006.

\bibitem{alam20}
M.~Alam, M.~M. Rahman, and J.~Widberg, ``Palmar: Towards adaptive
  multi-inhabitant activity recognition in point-cloud technology,'' \emph{IEEE
  International Conference on Computer Communications}, 2021.

\bibitem{benedek18}
C.~{Benedek}, B.~{Gálai}, B.~{Nagy}, and Z.~{Jankó}, ``Lidar-based gait
  analysis and activity recognition in a 4d surveillance system,'' \emph{IEEE
  Transactions on Circuits and Systems for Video Technology}, vol.~28, no.~1,
  pp. 101--113, 2018.

\bibitem{kingma2014adam}
D.~P. Kingma and J.~Ba, ``Adam: A method for stochastic optimization,''
  \emph{arXiv preprint arXiv:1412.6980}, 2014.

\bibitem{DBLP:journals/corr/SunS16a}
\BIBentryALTinterwordspacing
B.~Sun and K.~Saenko, ``Deep {CORAL:} correlation alignment for deep domain
  adaptation,'' \emph{CoRR}, vol. abs/1607.01719, 2016. [Online]. Available:
  \url{http://arxiv.org/abs/1607.01719}
\BIBentrySTDinterwordspacing

\bibitem{shu2018dirtt}
R.~Shu, H.~H. Bui, H.~Narui, and S.~Ermon, ``A dirt-t approach to unsupervised
  domain adaptation,'' 2018.

\bibitem{kim2020attract}
T.~Kim and C.~Kim, ``Attract, perturb, and explore: Learning a feature
  alignment network for semi-supervised domain adaptation,'' in \emph{European
  Conference on Computer Vision}.\hskip 1em plus 0.5em minus 0.4em\relax
  Springer, 2020, pp. 591--607.

\bibitem{kim2019highly}
D.~Kim, J.~Di~Carlo, B.~Katz, G.~Bledt, and S.~Kim, ``Highly dynamic quadruped
  locomotion via whole-body impulse control and model predictive control,''
  \emph{arXiv preprint arXiv:1909.06586}, 2019.

\bibitem{van2008visualizing}
L.~Van~der Maaten and G.~Hinton, ``Visualizing data using t-sne.''
  \emph{Journal of machine learning research}, vol.~9, no.~11, 2008.

\end{thebibliography}

\end{document}